\documentclass[journal]{IEEEtran}
\usepackage{cite}

\ifCLASSINFOpdf

\else

\fi

\usepackage{array}

\usepackage[caption=false,font=footnotesize]{subfig}

\usepackage{times}
\usepackage{epsfig}
\usepackage{graphicx}
\usepackage{amsmath}
\usepackage{amssymb}
\usepackage{epstopdf}
\usepackage{multirow}
\usepackage{CJK}

	\usepackage{subfig}

	\usepackage{color}
	
\begin{document}
		
		%
		\title{SAR-NAS: Skeleton-based Action Recognition via Neural Architecture Searching}
		%
		%
		%
		
		\author{Haoyuan Zhang,
			Yonghong Hou,~\IEEEmembership{Member,~IEEE,}
			Pichao Wang,~\IEEEmembership{Member,~IEEE,}
			Zihui Guo,\\
			Wanqing Li,~\IEEEmembership{Senior Member,~IEEE,}
			\thanks{This paper was supported by National Natural Science Foundation of China (61571325) in Tianjin, China. (Corresponding author: Pichao Wang)}
			\thanks{H. Zhang, Y. Hou and Z. Guo are with the School of Electronic Information Engineering, Tianjing University, Tianjin, China. (e-mail: zhy0860@tju.edu.cn; houroy@tju.edu.cn; gzihui@tju.edu.cn).}
			\thanks{P. Wang is with the Alibaba DAMO Academy, Bellevue, USA.
				(e-mail: pichaowang@gmail.com).}
			\thanks{W. Li is with the
				Advanced Multimedia Research Lab, University of Wollongong, Wollongong,
				Australia.
				(e-mail: wanqing@uow.edu.au).}
			
		}

		%
		%

		\markboth{}%
		{Shell \MakeLowercase{\textit{et al.}}:SAR-NAS:Skeleton-based Action Recognition via Neural Architecture Searching}
		%



		\maketitle
		
		\IEEEpubid{\begin{minipage}{\textwidth}\ \\ \\ \\ \\[8pt] \centering
			\end{minipage}}
			\begin{abstract}
					This paper presents a study of automatic design of neural network architectures for skeleton-based action recognition. Specifically, we encode a skeleton-based action instance into a tensor and carefully define a set of operations to build two types of network cells: normal cells and reduction cells. The recently developed DARTS (Differentiable Architecture Search) is adopted to search for an effective network architecture that is built upon the two types of cells.  All operations are 2D based in order to reduce the overall computation and search space. Experiments on the challenging NTU RGB+D and Kinectics datasets have verified that most of the networks developed to date for skeleton-based action recognition are likely not compact and efficient. The proposed method provides an approach to search for such a compact network that is able to achieve comparative or even better performance than the state-of-the-art methods.
			\end{abstract}
			
			\begin{IEEEkeywords}
				Neural Architecture Search, Action Recognition, Skeleton.
			\end{IEEEkeywords}

			%
			\IEEEpeerreviewmaketitle

			\section{Introduction}
			%
			%
			%
			%
			\IEEEPARstart{A}{s} an active  and challenging research topic in computer vision, human action recognition has many potential applications such as human computer interaction, autonomous retail shops, autonomous driving and somatic games~\cite{wang2018rgb,berretti2018representation}. In general, human action can be recognized from multiple modalities, such as RGB, optical flows, and skeletons. In the past decades, researcher mainly focused on RGB-based action recognition. With the advent of deep learning, neural networks are widely employed 
			to learn robust feature representations~\cite{karpathy2014large,tran2015learning,wang2018appearance,tran2018closer,wang2018non,feichtenhofer2019slowfast}. Although the RGB-based methods do well in the extraction of spatial feature form videos, motion between video frames are not well captured~\cite{simonyan2014two}, as a result optical flow is used for temporal complementation~\cite{wang2016temporal,feichtenhofer2017spatiotemporal}. Driven by the advances in skeleton estimation from RGB and depth modality, skeleton data is becoming a common modality in action recognition ~\cite{wang2015action,shi2017learning,wang2018depth,liu20173d,wang2017scene,wang2018cooperative,xiao2019action}. On the one hand, skeleton data is robust against the complex conditions such as occlusion, self-occlusion and variations of the subjects and viewpoints. On the other hand, skeleton data can be easily encoded into images so that conventional convolutional neural networks (CNN)~\cite{wang2016action,li2017joint,ke2017new,hou2016skeleton} and recurrent neural networks (RNN) \cite{du2015hierarchical,wang2017modeling,liu2017skeleton,li2018independently,si2019attention} can be employed. To date, many deep neural networks have been developed for skeleton-based action recognition \cite{tang2018deep,li2018co,wang2016action,song2017end}. However, design and tuning of such networks requires much time and human efforts. Moreover, the network architecture becomes more and more complicated in order to effectively exploit spatial-temporal information, hence more expensive to design and train~\cite{wang2018rgb}. Neural architecture search (NAS) has recently emerged as an effective approach to automating the design of a deep neural network. The key idea of NAS is to build a search space consisting of as many feasible network architectures as possible, then to explore the search space with an efficient algorithm and find out an optimal architecture for a given task under some constraints~\cite{elsken2019neural}. NAS has been successfully used to discover neural architectures for tasks such as image classification~\cite{zoph2018learning}, semantic image segmentation~\cite{liu2019auto} and object detection tasks~\cite{Ghiasi_2019_CVPR}. However, little work has been reported on discovering a suitable architecture for skeleton-based action recognition.

			This paper targets to fill this gap by exploiting NAS for skeleton-based action recognition. However, finding a feasible architecture in the large search space is usually time-consuming. The recent one-shot approach successfully reduces the cost of architecture discovery by an order of magnitudes compared with previous methods~\cite{liu2018darts,pham2018efficient}. DARTS (Differentiable Architecture Search)~\cite{liu2018darts} offers an alternative efficient method by relaxing the categorical choice of a particular operation in NAS to a softmax over all possible operations to make the search space continuous so that classic gradient-based optimization can be used. To take advantages of DARTS~\cite{liu2018darts}, we propose to adopt it to construct a compact CNN architecture for skeleton-based action recognition. An action instance is uniformly sampled into $T\times N \times 3$ data representation as input to the network, where $T$, $N$ are the number of frames and number of joints respectively and the 3D coordinates are treated as three channels. In order to construct an effective compact CNN architecture and also keep the search space of DARTS small, a set of operators is carefully chosen. First, the kernel sizes of all operators are relatively small and dilated convolution is used to increase the reception filed with less number of parameters. Second, squeeze-and-excitation module~\cite{hu2018squeeze} is chosen as an operator to capture the importance of each channel (i.e. motion in the Euclidean planes) to provide a channel-wise attention mechanism. Third, all operators are 2D-based for its low computational cost and high searching efficiency so that direct search on a large scale dataset without proxy is achieved. In addition, the 2D convolutional operations work along the temporal domain forming an effective Temporal 2D Convolutional Network (TCN). Through the multiple stages of convolution and pooling operations, local temporal information is gradually encoded into global one. The efficacy of these design strategies have been verified on two large scale datasets including NTU RGB+D~\cite{shahroudy2016ntu} and Kinetics~\cite{kay2017kinetics}. 
			
			\begin{figure*}
				\centering
				\includegraphics[height = 90mm, width = 170mm]{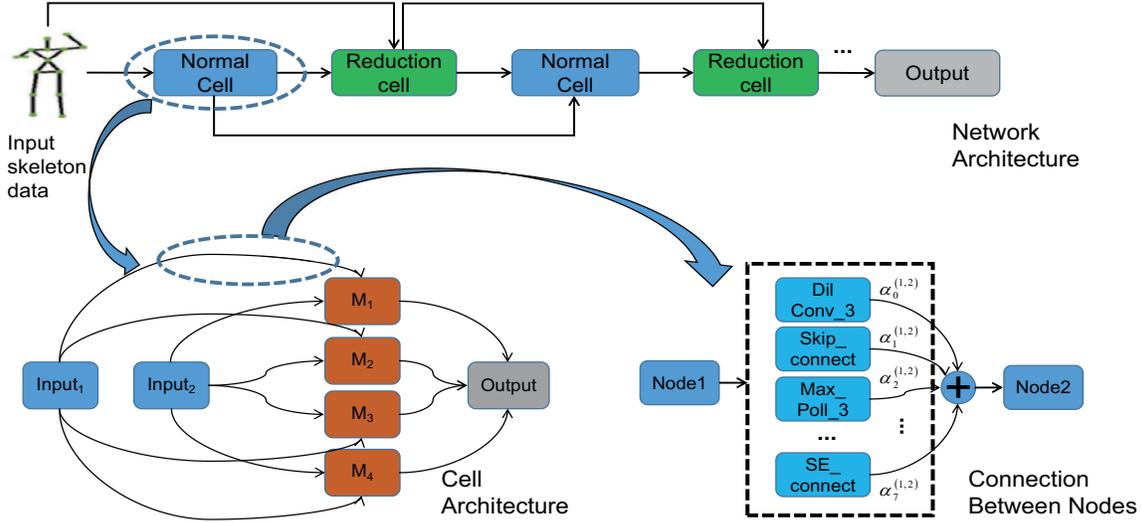}
				\caption{Overview of the architecture searching process. The network is stacked by several cells, and each cell takes the output of two previous cells as input. In each cell three are 7 nodes including two input nodes, four intermediate nodes and one output node. On each edge, a mixture of candidate operations is placed to relax the search space.}
				\label{fig:overview}
			\end{figure*}

			\section{Proposed Method}
			\label{sec:proposed_method}
			Fig.~\ref{fig:overview} shows the modularized network architecture $C$ to be searched. Follow the idea in DARTS~\cite{liu2018darts}, computation cells as the building blocks are defined and searched. The learned cells are then stacked together to form a complete network. Noted that a relative shallow network architecture is adopted since the limitation of GPU memory though the idea presented in this paper can be easily extended to a deep network.

			There are two types of cells, normal cells and reduction cells. The normal cells employ operations with stride 1 and reduction cells employ operations with stride 2 in order to halve the spatial dimension of the input. Each cell consists of several nodes and each node is a specific tensor-like feature map. A directed edge of two nodes works as an operation sampled from a set $O$ of candidate operators to transform one node to another. There are three types of nodes in a cell: input node, intermediate node and output node. Output tensors from previous cells are taken as input nodes, then through applying sampled operations to input nodes the intermediate nodes are generated, finally all intermediate nodes are concatenated to get the output node of current cell. Each cell is assumed to have two input nodes, four intermediate nodes and one output node, where 14 connections exit.

			As one of the most important elements in neural architecture search (NAS), search space is determined by the operators that may constitute an architecture and should be minimized for efficacy. Table~\ref{searchspace} lists the eight candidate operators that are carefully chosen for $O$. First, use of $3\times3$ dilated convolution (with dilation rate 2) and separable convolution as those in DARTS~\cite{liu2018darts}, removal of two $5\times5$ convolution operators, and inclusion of a normal $3\times3$ operator aim to reduce the number of parameters while keeping the same possible receptive field. Second, to use the squeeze-and-excitation module as a candidate operator aims to provide a channel-wise attention mechanism to select important channels. Third, as in DARTS~\cite{liu2018darts}, the max and average pooling operators provide two different ways to aggregate features, and skip-connect avoids vanishing gradient in deep architectures. Besides, for the sake of computational and searching efficiency, all operators are 2D based. For normal cells, all operations are of stride one and the convolved feature maps are padded to preserve their spatial resolution. For reduction cells located at the $\frac{1}{3}$ and $\frac{2}{3}$ of the total depth of a candidate network, all the operations adjacent to the input nodes are of stride two. The order of ReLU-Conv-BN is kept for convolutional operations, and each separable convolution is applied twice.

			\begin{table} 
				\centering
				\caption{Candidate operators.}\label{searchspace}
				\begin{tabular}{|c|c|}
					\hline
					Operators & Description\\
					\hline
					Conv3&Relu-Conv-Bn block with kernel size 3\\
					\hline
					SpeConv3&$3\times3$ separable convolutions\\
					\hline
					DilConv3&$3\times3$ dilated separable convolution \\
					\hline
					MaxPoll3&$3\times3$ max pooling \\
					\hline
					AveragePoll3&$3\times3$ average pooling \\
					\hline
					SkipConnect&Skip connection between nodes\\
					\hline
					SeConncet&Squeeze-and-excitation block\\
					\hline
					Zero&No connections\\
					\hline
				\end{tabular}
			\end{table}

			DARTS~\cite{liu2018darts} is adopted to search for cells and a network architecture with optimal performance. Specifically, the categorical choice of one certain operation is relaxed to a softmax over all possible operations, thus the task of cell architecture search is reduced to learning a set of continuous variables $\alpha = \{\alpha^{(i,j)}\}$ instead of directly searching the discrete architectures, here $(i,j)$ is the connection between node $i$ and $j$ and $\alpha^{(i,j)}$ means operation mixing weights for a pair of nodes $(i,j)$. Through this step the architecture is therefore encoded with $(\alpha_{normal},\alpha_{reduce})$, where $\alpha_{normal}$ and $\alpha_{reduce}$ are shared by all the normal cells and reduction cells, respectively. Then all operators in $O$ are introduced into each connection and modeled with architecture parameters. For each connection element-wise addition, referred to as mixed operation, is conducted between the outputs of each weighted operator.

			To jointly learn the architecture parameter $\alpha$ and the weights $\omega$ within all the mixed operations, DARTS~\cite{liu2018darts} treats the learning as a bi-level optimization problem, where $\alpha$ and $\omega$ are upper-level variable and lower-level variable, respectively. In particular, the weights $\omega$ of network is updated by minimizing training loss $L_{train}$ while architecture parameters are fixed, and the architecture parameter $\alpha$ is updated by minimizing validation loss $L_{val}$ while the weights of the network is fixed. This optimization process is expressed in Eqs. (1)\&(2),
			
			\begin{equation}
			\alpha^{*}=\mathop{\arg\min}_{\alpha}L_{val}{(\omega^{'},\alpha)},       \ \
			\end{equation}
			\begin{equation}
			\omega^{'}=\omega-\varepsilon\bigtriangledown_{\omega}L_{train}{(\omega,\alpha)}       \ \
			\end{equation}
			
			Here $\varepsilon$ is a small value for inner optimization. $\alpha$ is then updated with a gradient descent method with respect to the architecture parameters,
			\begin{equation}
			\alpha=\alpha-\gamma {\nabla _\alpha}, \ \
			\end{equation}
			\begin{equation}
			{\nabla _\alpha } = {\nabla _\alpha }{L_{val}}\left( {\omega ',\alpha } \right) - \varepsilon \nabla _{\alpha ,\omega }^2{L_{train}}\left( {\omega ,\alpha } \right){\nabla _{\omega '}}{L_{val}}\left( {\omega ',\alpha } \right), \ \
			\end{equation}
			where $\gamma$ is the learning rate. It is worth mentioning that within the optimization there exists a second-order derivative, which causes expensive computational costs, thus we adopt Hessian-vector products here to approximate the second-order derivative \cite{liu2018darts}, which can significantly reduce the computational complexity.

			\section{Experiments}
			\label{sec:results}
			\textbf{Dataset}: To evaluate efficacy of the proposed methods, comparative experiments were conducted on the widely used large NTU RGB+D \cite{shahroudy2016ntu} and Kinetics\cite{kay2017kinetics}.
			As one of the largest indoor-captured datasets for human action recognition, NTU RGB+D is currently widely used. It has in total 56000 action clips performed by 40 different actors. They are captured at different heights and horizontal angles in an indoor lab environment, with three cameras recording simultaneously. The actions cover 60 classes. Classes 1 to 49 are single-actor actions and classes 50 to 60 are actions involving two actors. The modalities of NTU RGB+D dataset include 3D skeleton, depth sequences, RGB and infrared videos. In our experiment only skeleton data were used. There are 25 joints for each subject and each joint is represented by its 3D location (X; Y; Z) in the camera coordinate system. There are two commonly used evaluation protocols: 
			\begin{itemize}
				\item Cross-subject (CS) in which there are 40320 training clips and 16560 clips for evaluation. In this setting, the training parts are from a subset of actors while the evaluating parts come from the rest of actors; 
				\item Cross-view (CV) in which there are 37920 clips for training and 18960 clips for evaluation. The training clips in this setting come from the second and third cameras and the clips captured by the first camera are for evaluation. We follow these protocols and report their top-1 recognition accuracy.
			\end{itemize}
			
			Deepmind Kinetics human action dataset \cite{kay2017kinetics} contains around 300, 000 video clips retrieved from YouTube, which covers 400 human action classes. The estimated joint locations obtained from OpenPose toolbox are used as skeleton data. The toolbox gives 2D coordinates (X, Y) in the pixel coordinate system and confidence scores C for the 18 joints. Then each joint is represented with a tuple of (X, Y, C) and a skeleton frame is recorded as an array of 18 tuples. Two persons are selected with the highest average joint confidence in each clip for the multi-person cases. In this way, one clip with T frames can be transformed into a skeleton sequence of these tuples. In the experiment, we use 240435 clips for training and 19796 clips for evaluation, which follows the AS-GCN ~\cite{li2019actional}. The top-1 and top-5 recognition accuracy are reported.

			\textbf{Implementation}: In the process of searching for an optimal architecture, network training and architecture update are alternatively performed. Each skeleton sequence of an action instance is preprocessed to $T=112$ frames through uniformly sampling and $N=50 joints$ in NTU RGB+D, $N=36 joints$ in Kinetics through padding with zeros if there is only one subject performing the action. The network in searching process is stacked by $6$ cells including 4 identical normal cells and 2 identical reduction cells searched by the method. Both two types of cells are initialized with 16 channels and consist of 7 nodes including 2 input, 4 intermediate and 1 output nodes. We search for 35 epochs, initial learning rate is set as 0.015 and is decreased at each iteration. When a search procedure is finished, an optimal $6$ cell network is returned according to the learned architecture parameters and the network is trained from scratch. In the network, two top-weighted non-zero operations are retained. 
			
			\begin{figure}
				\centering
				{\includegraphics[height = 40mm, width = 35mm]{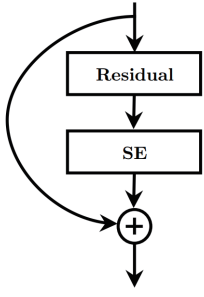}} 
				\caption{Standard SE-ResNet block~\cite{hu2018squeeze}.}\label{fig2}
				\label{fig:cs}
			\end{figure}
			
			\begin{figure}
				\centering
				{\includegraphics[height = 71mm, width = 80mm]{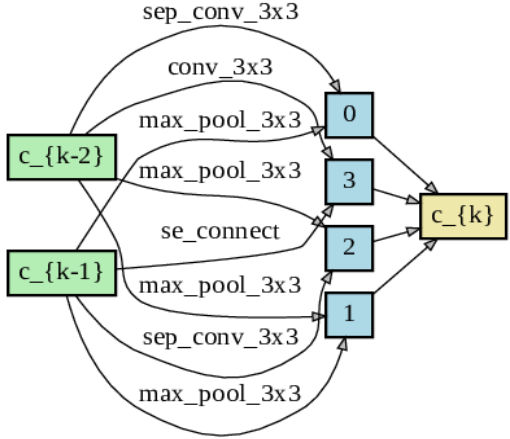}} 
				\caption{Normal cell structure searched by the proposed method.}\label{fig3}
				\label{fig:ca}
			\end{figure}
			
			\begin{figure}
				\centering
				{\includegraphics[height = 43mm, width = 88mm]{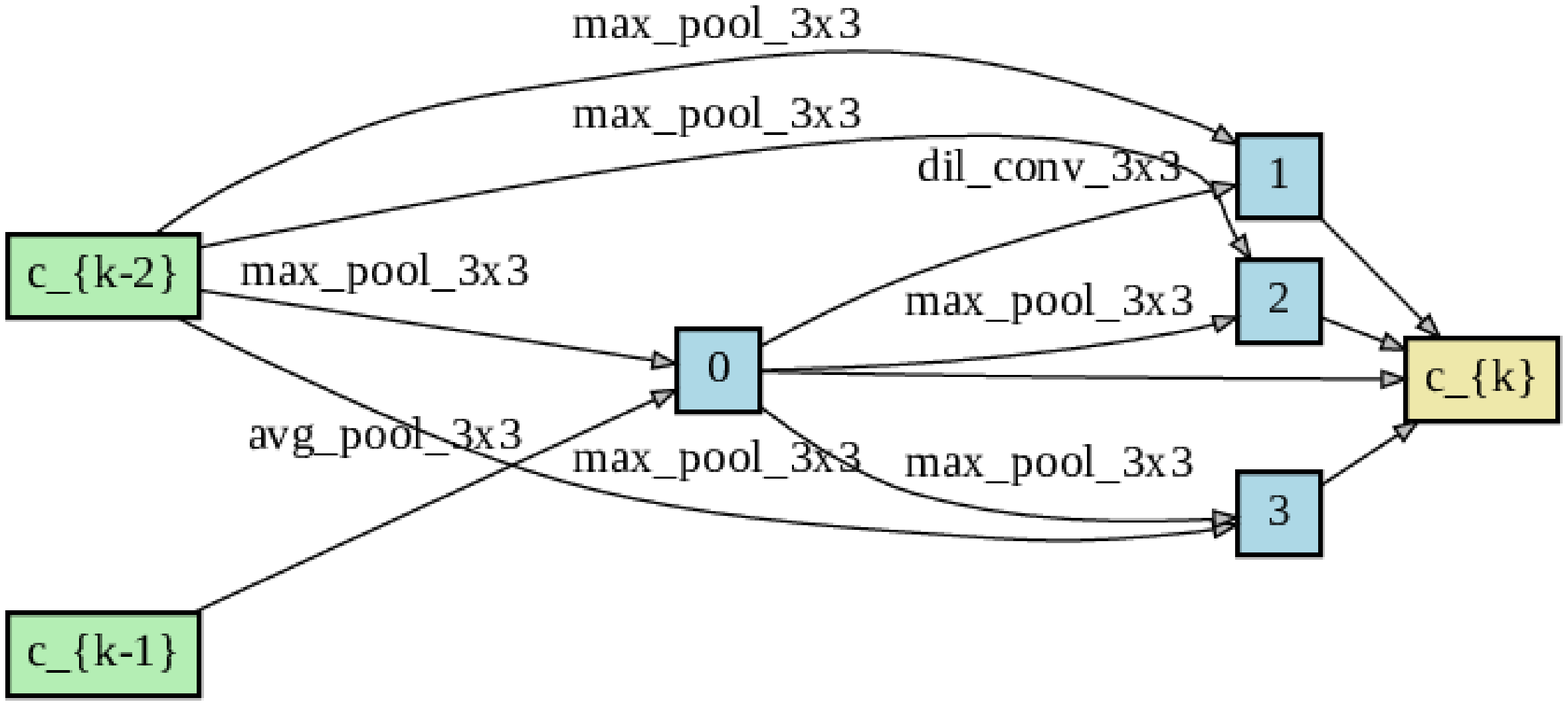}}
				\caption{Reduction cell structure searched by the proposed method.}\label{fig4}
				\label{fig:cv}
			\end{figure}

			The final network is built using the searched optimal cell structures. Based on the performance on NTU RGB+D using the cross-subject protocol shown in Table~\ref{tab2}, it is found that the network stacked by $9$ cells achieved the best performance. Noted that the reduction cells locate at the third and sixth layers of the network. Then the final network is trained with 350 epochs, the initial learning rate is set as 0.01 and decreased each iteration steps with a momentum of 0.9.
			\begin{table}
				\centering
				\caption{Performance comparison of final network with different depth of cells on the NTU RGB+D dataset using cross-subject protocol.}\label{tab2}
				\begin{tabular}{|c|c|c|}
					\hline
					Depth of cells &Top-1 accuracy&Flops\\
					\hline
					7&81.2\%&8.09G\\
					\hline
					8&84.1\%&9.09G \\
					\hline
					9&86.4\%&10.2G\\
					\hline
					10&85.0\%&11.1G\\
					\hline
					11&84.8\%&12.1G\\
					\hline
				\end{tabular}
			\end{table}
			
			\begin{table}
				\centering
				\caption{Performance comparison with the state-of-the-art methods on the NTU RGB+D dataset using cross-subject and cross-view protocol.}\label{tab3}
				\begin{tabular}{|c|c|c|c|}
					\hline
					Architecture&Parameters&cross-subject&cross-view\\
					\hline
					LSTM-CNN~\cite{li2017skeleton}&60M+&82.9\%&90.1\%\\
					\hline
					DCM (DenseNet161)~\cite{xiao2019self}&10M+&84.5\%&91.3\% \\
					\hline
					DPRL + GCNN~\cite{tang2018deep}&-&83.5\%&89.8\%\\
					\hline
					HCN~\cite{li2018co}&2.64M&86.5\%&91.1\%\\
					\hline
					ST-GCN~\cite{yan2018spatial}&3.1M&81.5\%&88.3\%\\
					\hline
					AS-GCN~\cite{li2019actional}&7.1M&86.8\%&94.2\%\\
					\hline
					DARTS~\cite{liu2018darts}&1.6M&83.9\%&92.0\%\\
					\hline
					SAR-NAS (Ours)&1.3M&86.4\%&94.3\%\\
					\hline
				\end{tabular}
			\end{table}
			
			\begin{table}
				\centering
				\caption{Performance comparison with the state-of-the-art methods on the Kinetics dataset }\label{tab4}
				\begin{tabular}{|c|c|c|c|}
					\hline
					Architecture&Parameters&Top-1 accuracy&Top-5 accuracy\\
					\hline
					Temporal Conv~\cite{kim2017interpretable}&-&20.3\%&40.0\%\\
					\hline
					ST-GCN~\cite{yan2018spatial}&3.1M&30.7\%&52.8\%\\
					\hline
					AS-GCN~\cite{li2019actional}&7.1M&34.8\%&56.5\%\\
					\hline			
					DARTS~\cite{liu2018darts}&2.7M&32.1\%&54.0\%\\
					\hline
					SAR-NAS (Ours)&2.5M&33.6\%&56.3\%\\
					\hline
				\end{tabular}
			\end{table}
			
			\begin{table}
				\centering
				\caption{Comparison of the recognition of actions with and without including the SE module as an candidate operation on the NTU RGB+D dataset using cross-view protocol.}\label{tab5}
				\begin{tabular}{|c|c|c|}
					\hline
					Actions&Accuracy with SE&Accuracy without SE\\
					\hline
					Drink water&95.0\%&94.1\% \\
					\hline
					Brush hair&94.6\%&91.5\% \\
					\hline
					Wipe face&88.3\%&84.5\% \\
					\hline
					Back pain&89.2\%&87.0\% \\
					\hline
					Kicking&97.1\%&95.9\% \\
					\hline
					Touch pocket&95.3\%&91.1\% \\
					\hline
				\end{tabular}
			\end{table}
			
			\textbf{Results}: The search was performed on a single NVIDIA TitanXP GPU and the total search time on the NTU RGB+D dataset is about 29 hours.  Fig.~\ref{fig3} and Fig.~\ref{fig4} show the searched normal and reduction cell architectures. The final network with $9$ cells is established and trained for 350 epochs. Top-1 accuracy on the NTU RGB+D dataset obtained by the $9$ cell network and its comparison to the state-of-the-art results are shown in Table~\ref{tab3}. Besides, the number of parameters of different network architectures are shown in the Table as well. The results have demonstrated that the proposed method achieved superior performance with much fewer parameters than any other methods. In particular, compared with the most advanced GCN method \cite{li2019actional} and CNN-based method \cite{li2018co} on cross-subject test, the number of parameters of the proposed method is decreased by 81.7\% and 58.1\%, respectively, which shows that the searched network architecture is much more compact. On Kinetics, recognition performances of different architectures in terms of top-1 and top-5 accuracies are compared in Table~\ref{tab4}. The proposed method achieves competitive performance compared with the state-of-the-art. Due to the noisy skeleton data and the limitation of CNN itself to process skeleton data compared with GCN, in terms of top-1 accuracy the proposed method is slightly lower than AS-GCN. However, the number of parameters is decreased by 64.8\%, which shows the high efficiency of proposed method.

			\textbf{Ablation study}: To validate the effectiveness of the chosen operations and hence the search space, experiments are also conducted using the operations defined in~\cite{liu2018darts} to search for an optimal network for the NTU RGB+D dataset using cross-subject and cross-view protocols. Table~\ref{tab3} and Table~\ref{tab4} show the results. It is observed that the network search from the operations defined in this paper has improved the accuracy by $2.5$ and $2.3$ percentage points on cross-subject and cross-view tasks, respectively, and the number of parameters reduced by 0.3M.
			
			To verify the effectiveness of the SE module, performances of two searched networks with and without the SE operation is compared,  Table~\ref{tab5} shows the recognition accuracy of some representative actions. The results have demonstrated that the SE operation is useful as expected. Moreover, visually compared with the hand-designed SE-ResNet~\cite{hu2018squeeze} architecture for image classification as shown in Fig.~\ref{fig2}, the position of SE module of searched network architecture is different,  with the hand-designed SE module placed in the end of each residual block while the searched one located in the middle of the normal cell, as shown in Fig.~\ref{fig3}.

			\section{Conclusion}
			\label{sec:conclusion_future_work}
			
			This paper presents the first study on searching for a network architecture for skeleton-based action recognition. The results on the NTU RGB+D dataset have shown that most of the networks developed to date for skeleton-based action recognition are likely not compact and efficient. The proposed method provides an approach to searching for such a compact network that is able to achieve comparative or even better performance than the state-of-the-art methods though the searched network architecture may be dataset-dependent.

			\bibliography{jvci_manuscript}

		\end{document}